\documentclass[10pt,twocolumn,letterpaper]{article}

\usepackage{wacv}
\usepackage{times}
\usepackage{epsfig}
\usepackage{graphicx}
\usepackage{amsmath}
\usepackage{amssymb}
\usepackage{subfiles}
\usepackage{booktabs}
\usepackage{caption}
\usepackage{subcaption}
\usepackage{placeins}
\usepackage{array,multirow}
\usepackage[title]{appendix}
\usepackage{breqn}
\usepackage{pgfplots}
\usepackage{tikz}
\usepgfplotslibrary{colorbrewer}
\pgfplotsset{cycle list/Set1-9}
\tikzset{every picture/.style={line width=1pt}}
\usetikzlibrary{patterns}

\newcommand{\C}{\mathcal{C}}
\newcommand{\Cbase}{\mathcal{C}_{\text{base}}}
\newcommand{\Cnovel}{\mathcal{C}_{\text{novel}}}
\newcommand{\SData}{\mathcal{S}}
\newcommand{\Strain}{\mathcal{S}_{\text{train}}}
\newcommand{\Stest}{\mathcal{S}_{\text{test}}}
\newcommand{\Strainnovel}{\mathcal{S}_{\text{train}}^{\text{novel}}}
\newcommand{\Straink}{\mathcal{S}_{\text{train}}^{\text{k}}}

\DeclareMathOperator*{\argmin}{arg\,min}

\wacvfinalcopy 
\def\wacvPaperID{1257} 

\ifwacvfinal\fi
\begin{document}

\title{Multimodal Prototypical Networks for Few-shot Learning}

\newcommand*\samethanks[1][\value{footnote}]{\footnotemark[#1]}

\makeatletter
\def\@fnsymbol#1{\ensuremath{\ifcase#1\or \dagger\or \ddagger\or
   \mathsection\or \mathparagraph\or \|\or **\or \dagger\dagger
   \or \ddagger\ddagger \else\@ctrerr\fi}}

\author{Frederik Pahde\textsuperscript{1, }\thanks{Work completed while at SAP AI Research, prior to joining Amazon.com, Inc.} \ , Mihai  Puscas\textsuperscript{2, }\thanks{Work completed while at SAP AI Research, prior to joining Huawei Research, Ireland} \ , Tassilo Klein\textsuperscript{3}, Moin Nabi\textsuperscript{3} \\
\textsuperscript{1}Amazon.com, Inc., \textsuperscript{2}Huawei Research, Ireland, \textsuperscript{3}SAP AI Research, Berlin, Germany\\
{\tt\small frederikpahde@gmail.com, mihai.puscas@huawei.com,  \{tassilo.klein, m.nabi\}@sap.com}
}

\maketitle
\ifwacvfinal\thispagestyle{empty}\fi

\begin{abstract}
Although providing exceptional results for many computer vision tasks, state-of-the-art deep learning algorithms catastrophically struggle in low data scenarios. However, if data in additional modalities exist (e.g. text) this can compensate for the lack of data and improve the classification results. To overcome this data scarcity, we design a cross-modal feature generation framework capable of enriching the low populated embedding space in few-shot scenarios, leveraging data from the auxiliary modality. Specifically, we train a generative model that maps text data into the visual feature space to obtain more reliable prototypes. This allows to exploit data from additional modalities (e.g. text) during training while the ultimate task at test time remains classification with exclusively visual data. We show that in such cases nearest neighbor classification is a viable approach and outperform state-of-the-art single-modal and multimodal few-shot learning methods on the CUB-200 and Oxford-102 datasets.

\end{abstract}
\section{Introduction}
\label{sec:introduction}
Despite the great success of deep learning models, the necessity of large training sets for these models is often a limiting factor. Many applications come with the natural problem of limited data, making it too expensive or even impossible to collect a sufficient number of training samples and leading to poor model accuracies. This is in contrast to the human ability to quickly learn new concepts. Consequently, the study of few-shot classification, i.e. learning new concepts from a very limited amount of training data, has gathered focus in recent years~\cite{lake_one_2011, vinyals_matching_2016, bertinetto_learning_2016, ravi_optimization_2017, hariharan_low-shot_2017, snell_prototypical_2017, raza2019weakly}. Finetuning DNNs has been shown to be effective in a context where the big data assumption holds \cite{oquab2014learning}. However, scenarios where access is limited to only very few samples of novel data are extremely susceptible to over-fitting.\\
\begin{figure}[t]
	\centering
  \includegraphics[width=0.4\textwidth]{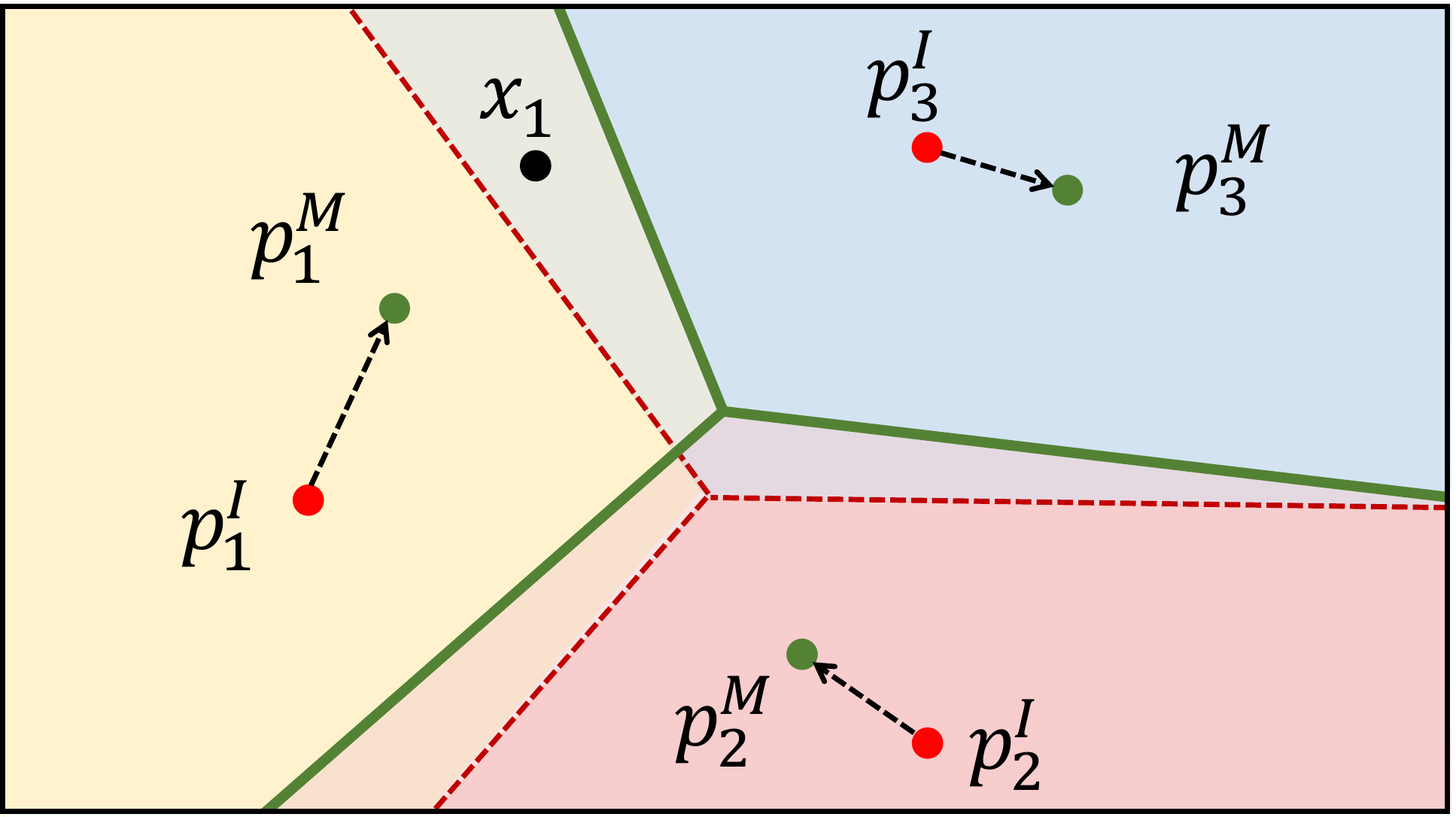}
	\caption{Multimodal Prototypical Networks: Cross-modal generated representations can condense the embedding space and move the visual prototypes $p^I$ towards more reliable multimodal prototypes $p^M$ and it improves the classification accuracy on unseen test samples (e.g., $x_1$).}
	\label{fig:taskFigure}
\end{figure}

Originally, few shot learning defined a scenario where the only very few samples per class were accessible \cite{li2006one, koch_siamese_2015, lake_one_2011}. With the advent of deep learning the assumption was broadened, into having large amounts of data accessible for a number of \textit{base} classes, with \textit{novel} classes bound by a scarce data regime. This more realistic scenario falls under a meta-learning context, where a representation is learned on the base classes to be employed later on the novel classes.

To leverage the powerful representations that can be learned on the base classes with a DNN, a wide variety of meta-learning methods have been proposed. Santoro et al.~\cite{santoro2016meta} make use of a memory network to better assimilate new data and make predictions using it. Edwards et al.~ \cite{edwards2016towards} aim to make use of learned dataset statistics to better fine-tune on new samples.\\
In contrast to more \emph{model-driven} methods, ~\cite{vinyals_matching_2016} learns an embedding of the labelled examples over which an attention mechanism can be utilized, while ~\cite{snell_prototypical_2017} learns a mapping from the input to an embedding for which its class is represented by a prototype. Upon learning an embedding, both methods make use of a simple k-nearest neighbor approach to infer the class membership of unseen samples, implying that they can leverage the representational power of DNNs in a low data regime. See also \cite{wang2019simpleshot} for a
discussion.\\
However, even when optimizing the learning process for the low-shot scenario, the lack of novel samples remains a hindrance. To mitigate this, a series of generative approaches have been developed, increasing the number of novel class samples that can be utilized during training. Hariharan et al.~\cite{hariharan_low-shot_2017} facilitates training the classifier by generating features, disregarding realism or diversity criteria. While this approach provides a stable meta-learning process, and practically generates useful hallucinated samples, the diversity of generated samples is bound by the samples used to learn the generator. 
The key assumption is that incorporating multimodal data can provide the means to inject diversity into the generation process. This is achieved by learning a cross-modal mapping which in turn broadens the scope of the generated sample space.
The most closely related work to us is \cite{pahdeWACV}, which makes use of additional textual data in an adversarial context, followed by a self-paced selection of the most discriminative samples.

Our method builds upon the observation that the representations learned through DNNs are powerful enough for the use of simple non-parametric classification techniques \cite{bauer2017discriminative}, and that multi-modal data can improve generation diversity.
To this end, an image encoder is first trained on the available base classes, after which a text-conditional GAN learns a cross-modal mapping between the textual and visual embedding spaces. This mapping can then be used to generate feature representations that reside in the visual space, conditioned by textual data. Intuitively, our method makes use of the cross-modal feature mapping to shift single-modal prototypes $p^I$ (representing visual data) to $p^M$, mimicking unseen samples of the novel classes. This process can be observed in Fig. \ref{fig:taskFigure}, where a given sample $x_i$ is classified differently though the shift in the prototypes. In a prototypical space, k-NN, a non-parametric classification technique is used, and thus only the representation learning stage requires multi-modal data, the inference stage requiring only visual data.


The main contributions of this work include the use of a cross-modal feature generation network in the context of few-shot learning. Furthermore, we suggest a strategy to combine real and generated features, allowing us to infer the class membership of unseen samples with a simple nearest neighbor approach. Our method outperforms our baselines and the state-of-the-art approaches for multimodal and image-only few-shot learning by a large margin for the CUB-200 and Oxford-102 datasets.


\section{Related Work}
In this section we briefly review the previous works related to few-shot learning and multimodal learning.

\subsection{Few-Shot Learning}
For learning deep networks using limited amounts of data, different approaches have been developed in recent years. Following Taigman et al.~\cite{taigman2014deepface}, Koch et al.~ \cite{koch_siamese_2015} interpreted this task as a verification problem, i.e. given two samples, it has to be verified, whether both samples belong to the same class. Therefore, they employed siamese neural networks \cite{bromley1994signature} to compute the distance between the two samples and perform nearest neighbor classification in the learned embedding space. Some recent works approach few-shot learning by striving to avoid overfitting by modifications to the loss function or the regularization term. Yoo et al.~\cite{yoo_efficient_2017} proposed a clustering of neurons on each layer of the network and calculated a single gradient for all members of a cluster during the training to prevent overfitting. The optimal number of clusters per layer is determined by a reinforcement learning algorithm. A more intuitive strategy is to approach few-shot learning on data-level, meaning that the performance of the model can be improved by collecting additional related data. Douze et al.~\cite{douze_low-shot_2017} proposed a semi-supervised approach in which a large unlabeled dataset containing similar images was included in addition to the original training set. This large collection of images was exploited to support label propagation in the few-shot learning scenario. Hariharan et al.~\cite{hariharan_low-shot_2017} combined both strategies (data-level and algorithm-level) by defining the squared gradient magnitude loss, that forces models to generalize well from only a few samples, on the one hand and generating new images by hallucinating features on the other hand. For the latter, they trained a model to find common transformations between existing images that can be applied to new images to generate new training data \cite{wang_low-shot_2018}. Other recent approaches to few-shot learning have leveraged meta-learning strategies. Ravi et al.~\cite{ravi_optimization_2017} trained a long short-term memory (LSTM) network as meta-learner that learns the exact optimization algorithm to train a learner neural network that performs the classification in a few-shot learning setting. This method was proposed due to the observation that the update function of standard optimization algorithms like SGD is similar to the update of the cell state of a LSTM. Similarly, Finn et al.~\cite{finn2017model} suggested a model-agnostic meta-learning approach (MAML) that learns a model on base classes during a meta learning phase optimized to perform well when finetuned on a small set of novel classes.
Moreover, Bertinetto et al.~\cite{bertinetto_learning_2016} trained a meta-learner feed-forward neural network that predicts the parameters of another, discriminative feed-forward neural network in a few-shot learning scenario. Another technique that has been applied successfully to few-shot learning recently is attention. \cite{vinyals_matching_2016} introduced matching networks for one-shot learning tasks. This network is able to apply an attention mechanism over embeddings of labeled samples in order to classify unlabeled samples. One further outcome of this work is that it is helpful to mimic the one-shot learning setting already during training by defining mini-batches, called few-shot episodes with subsampled classes. Snell et al.~\cite{snell_prototypical_2017} generalize this approach by proposing prototypical networks. Prototypical networks search for a non-linear embedding space (the prototype) in which classes can be represented as the mean of all corresponding samples. Classification is then performed by finding the closest prototype in the embedding space. Other related works include \cite{wertheimer2019few,liu2018learning,pahde2019low}. 

\subsection{Multimodal Learning}
Kiros et al~\cite{kiros_unifying_2014} propose to align visual and semantic information in a joint embedding space using a encoder-decoder pipeline to learn a multimodal representation.
Building upon this, Faghri et al~\cite{faghri_vse++:_2017} improve the mixed representation by incorporating a triplet ranking loss.\\
Karpathy et al~\cite{karpathy_deep_2015} generate textual image descriptions given the visual data. Their model infers latent alignments between regions of images and segments of sentences of their respective descriptions. 
Reed et al~\cite{reed_learning_2016} focus on fine-grained visual descriptions. 
They present an end-to-end trainable deep structured joint embedding trained on two datasets containing fine-grained visual descriptions.\\
In addition to multimodal embeddings, another related field using data from different modalities is text-to-image generation. 
Reed et al~\cite{reed16_gen} study image synthesis based on textual information. Zhang et al~\cite{zhang_stackgan++:_2017} greatly improve the quality of generated images to a photo-realistic high-resolution level by stacking multiple GANs (StackGANs). 
Extensions of StackGANs include an end-to-end trainable version \cite{zhang_stackgan++:_2017} and considering an attention mechanism over the textual input \cite{xu_attngan:_2017}. Sharma et al.~\cite{sharma2018chatpainter} extended the conditioning by involving dialogue data and further improved the image quality.
Beside the usage of GANs for conditioned image generation, other work employed Variational Autoencoders \cite{kingma2013auto} to generate images \cite{mishra2017generative}. However, they conditioned on attribute vectors instead of text.\\
Some works have leveraged multimodal data to improve classification results. Elhoseiny et al~\cite{elhoseiny_link_2017} collect noisy text descriptions and train a model that is able to connect relevant terms to its corresponding visual parts. This allows zero-shot classification for unseen samples, i.e. visual samples for novel classes do not exist. Similarly, Zhu et al.~\cite{zhu2018generative} train a classifier with images generated by a GAN given noisy text descriptions and test their approach in a zero-shot setup. Xian et al~\cite{xian2018feature} follow this notion, however, generating feature vectors instead of images. In the context of few-shot learnig, Pahde et al~\cite{pahdeWACV,pahde2018discriminative,pahde2018cross} have leveraged textual descriptions to generate additional training images, as opposed to visual feature embeddings generated in this work. Along with a self-paced learning strategy for sample selection this method improves few-shot learning accuracies. 

\section{Multimodal Prototypical Networks}
To define our developed method we first introduce the necessary notation and then  describe the architecture of our framework.
\begin{figure*}[t!]
	\centering
  \includegraphics[width=0.95\textwidth]{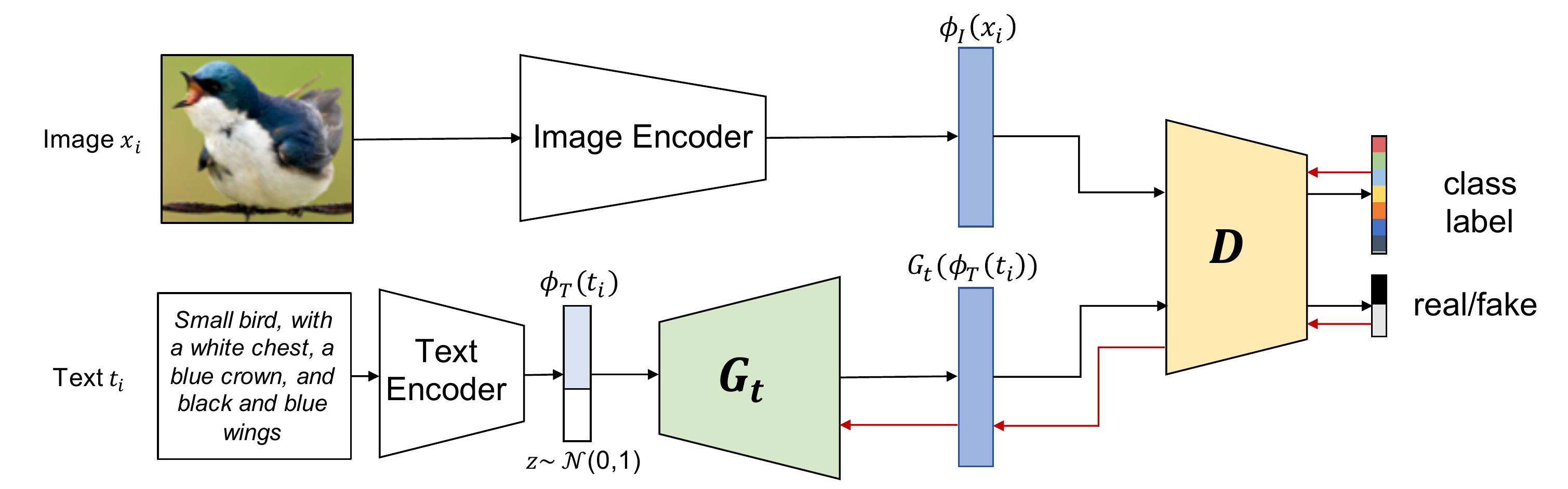}
	\caption{Architecture of our proposed feature-generating network for few-shot learning: The GAN framework containing a generator $G_t$ and a discriminator D is optimized to transform a text embedding given a pre-trained text encoder into the visual embedding space $\varphi$ yielded by a pre-trained image encoder. The discriminator computes a reconstruction loss (real/fake) and an auxiliary classification loss.}
	\label{fig:ourMethod}
\end{figure*}

\subsection{Preliminaries}
Let $\mathcal{I}$ denote the image space, $\mathcal{T}$ the text space and $\C=\lbrace 1,...,R\rbrace$ be the discrete label space. Further, let $x_i \in \mathcal{I}$ be the $i$-th input data point, $t_i \in \mathcal{T}$ its corresponding textual description and $y_i \in \C$ its label.
In the few-shot setting, we consider two disjunct subsets of the label space: $\Cbase$ -  labels for which we have access to sufficient  data samples, and  $\Cnovel$ novel classes, which are underrepresented in the data. Note that both subsets exhaust the label space $\C$, i.e. $\C = \Cbase \cup \Cnovel$. We further assume that in general $|\Cnovel| < |\Cbase|$.\\
We organize the data set $\SData$ as follows.
Training data $\Strain$ consists of tuples $\{(x_i, t_i, y_i)\}_{i=1}^{n}$ taken from the whole data set and test data $\Stest = \{(x_i, y_i) : y_i \in \Cnovel\}_{i=1}^m$ that belongs to novel classes such that $\SData = \Strain \cup \Stest$, $\Strain \cap \Stest = \emptyset$.
Naturally, we can also consider $\Strainnovel = \{(x_i, t_i, y_i) : (x_i, t_i, y_i) \in \Strain, y_i \in \Cnovel\}_{i=1}^k \subset \Strain$,
where in accordance with a few-shot scenario $k = \left|\Strainnovel\right|\ll\left|\Strain\right| = n$. 
Additionally, in a few-shot learning scenario, the number of samples per category of $\Cbase$ may be limited to $g$, denoted by $\Strainnovel(g)$. Note that contrary to the benchmark defined by Hariharan et al.~\cite{hariharan_low-shot_2017}, the few-shot learning scenario in this paper is multimodal in training. However, the testing phase is single-modal on image data of $\Cnovel$.

\subsection{Nearest Neighbor in Visual Embedding Space}
The classification in the embedding space is performed with a simple nearest neighbor approach. The assumption is that given a powerful feature representation, such as ResNet-18 feature vectors, nearest neighbor is a viable choice as classification model and has proven to outperform more sophisticated few-shot learning approaches \cite{bauer2017discriminative}. 
Therefore, we use the visual data from base classes $\Cbase$ to train an image encoder $\Phi_I$, providing a discriminative visual embedding space $\varphi$. For novel visual samples $x_i \in \Strainnovel$, $\Phi_I(x_i)$ then provides the embedding accordingly, featuring discriminativeness given by the pre-trained visual embedding space $\varphi$.\\
Following \cite{snell_prototypical_2017}, for every novel class $k\in\Cnovel$ we calculate a visual prototype $p^k$ of all encoded training samples:
\begin{equation}\label{eq:protVis}
    p^k = \frac{1}{|\Straink|}\sum_{(x_i,y_i) \in \Straink}\Phi_I(x_i),
\end{equation}
where $\Straink = \{(x_i,y_i) \in \Strainnovel | y_i=k\}_{i=1}^{n}$ is the set of all training pairs $(x_i,y_i)$ for class $k$. 
Classification of test samples is performed by finding the closest prototype given a distance function $d(\cdot)$. Thus, given a sample $x^{\text{test}} \in \Stest$ the class membership is predicted as follows:
\begin{equation}\label{eq:nn}
    c = \argmin_k d(\Phi_I(x_{\text{test}}), p^k)
\end{equation}
This assigns the class label of the closest prototype to an unseen test sample. Given the assumption that $\varphi$ is a discriminative representation of visual data, Eq. \ref{eq:nn} provides a powerful classification model. However, due to the few-shot scenario and the intrinsic feature sparsity in training space, $\Strainnovel$ is rather limited such that the computed class prototypes $\{p^k:k\in\Cnovel\}$ consequentially yields merely a rough approximation of the true class mean.

\begin{table*}[ht]
    \setlength{\tabcolsep}{2.5pt}
	\centering{
		\begin{tabular}{lcccccc} \toprule
		Dataset & Method  & 1-shot & 2-shot & 5-shot & 10-shot &20-shot \\ \midrule
		CUB & Pahde et al. \cite{pahdeWACV} & 57.67 & 59.83 & 73.01 & 78.10 & 84.24 \\
		& Image Only Baseline(Resnet-18+NN) & 62.65$\pm$0.22 & 73.52$\pm$0.15 & 82.44$\pm$0.09 & 85.64$\pm$0.08 & 87.27$\pm$0.08\\
		& ZSL Baseline (Generated Resnet-18+NN) & 58.28$\pm$0.22 & 65.62$\pm$0.19 & 71.79$\pm$0.14 & 74.15$\pm$0.11 & 75.32$\pm$0.13 \\
		& Our Method (Multimodal Resnet-18 + NN) & \textbf{70.39$\pm$0.19} & \textbf{78.62$\pm$0.12} & \textbf{84.32$\pm$0.06} & \textbf{86.23$\pm$0.08} & \textbf{87.47$\pm$0.09}\\ \midrule
		Oxford-102 & Pahde et al. \cite{pahdeWACV} & 78.37 & 91.18 & 92.21 & - & - \\
		& Image Only Baseline (Resnet-18+NN) & 84.18$\pm$0.48 &  90.25$\pm$0.20 & 94.18$\pm$0.13 & 95.63$\pm$0.14 & 96.25$\pm$0.10\\
		& ZSL Baseline (Generated Resnet-18+NN) & 73.35$\pm$0.52 & 77.52$\pm$0.34 & 81.14$\pm$0.25 & 82.95$\pm$0.28 & 83.97$\pm$0.21 \\
		& Our Method (Multimodal Resnet-18 + NN) & \textbf{86.52$\pm$0.36}& \textbf{91.31$\pm$0.18} & \textbf{94.57$\pm$0.13} & \textbf{95.74$\pm$0.13} & \textbf{96.38$\pm$0.10}\\\bottomrule
		\end{tabular}
	}
	\captionsetup{position=above}
	\captionof{table}{Top-5 accuracy in comparison to other multimodal few-shot learning approaches and our baselines for CUB-200 (50-way classification) and Oxford-102 (20-way classification) datasets with $n\in\{1,2,5,10,20\}$}
	\label{tab:results_multimodal}
\vspace{-0.5cm}

\end{table*}

\subsection{Cross-modal Feature Generation}
A viable solution to enrich the training space to enable the calculation of more reliable estimations of the class prototypes is to leverage the multimodality in $\Strainnovel$. Thus, the core idea of our method is to use textual descriptions provided in the training data to generate additional visual feature vectors compensating the few-shot feature sparsity. Therefore, we propose to train a text-conditional generative network $G_t$ that learns a mapping from the encoded textual description into the pre-trained visual feature space $\varphi$ for a given training tuple $(x_i,t_i,y_i)$ according to
\begin{equation}
    G_t(\Phi_T(t_i)) \approx \Phi_I(x_i).
\end{equation}
For the purpose of cross-modal feature generation we use a modified version of text-condtional generative adversarial networks (tcGAN) \cite{reed16_gen, zhang_stackgan++:_2017, xu_attngan:_2017}. The goal of tcGAN is to generate an image given its textual description in the GAN framework~\cite{goodfellow_generative_2014}. More specifically, the tcGAN is provided with an embedding $\phi_T(\cdot)$ of the textual description. 
Therefore, a common strategy is to define two agents $G$ and $D$ solving the adversarial game of generating images that cannot be distinguished from real samples ($G$) and detecting the generated images as fake ($D$). Because our strategy is to perform nearest-neighbor classification in a pre-trained embedding space $\varphi$, we slightly change the purpose of tcGAN. Instead of generating images $x_i \in \mathcal{I}$, we optimize $G$ to generate its feature representation $\Phi_I(x_i)$ in the space $\varphi$. Generally, the representation vector in an embedding space has a significantly lower dimensionality than the original image. Consequentially, the generation of feature vectors is a computational cheaper task compared to the generation of images, can be trained more efficiently and is less error-prone.\\
To this end, using data from $\Strain$ our modified tcGAN can be trained by optimizing the following loss function,
\begin{dmath}
\label{tcGANloss}
\mathcal{L}_{tcGAN}\left(G_t,D\right)=\mathbb{E}_{x_i\sim p_{data}}\left[\log D\left(\Phi_I(x_i)\right)\right]+\mathbb{E}_{t_i\sim p_{data},z}\left[\log D\left(G_t\left(\Phi_T(t_i),z\right)\right)\right],
\end{dmath}
which entails the reconstruction loss that is used for the traditional GAN implementation \cite{goodfellow_generative_2014}. 
Moreover, following \cite{odena2016conditional, pahdeWACV, zhu2018generative} we define the auxiliary task of class prediction during the training of the tcGAN. This entails augmenting the tcGAN loss given in Eq.~\ref{tcGANloss} with a discriminative classification term, which is defined as
\begin{equation}
\mathcal{L}_{class}\left(D\right)=\mathbb{E}_{C,I}\left[\log p\left(C\mid I\right)\right]
\end{equation}
\begin{equation}
\text{and} \quad \mathcal{L}_{class}\left(G_t\right)\triangleq\mathcal{L}_{class}\left(D\right).
\end{equation}
Augmenting the original GAN loss with the defined auxiliary term, the optimization objectives for $D$ and $G_t$ can now be defined as
\begin{equation}\mathcal{L}\left(D\right)=\mathcal{L}_{tcGAN}\left(G_t,D\right)+\mathcal{L}_{class}\left(D\right)
\end{equation}
\begin{equation}\mathcal{L}\left(G_t\right)=\mathcal{L}_{tcGAN}\left(G_t,D\right)-\mathcal{L}_{class}\left(G_t\right),
\end{equation}
which are optimized in an adversarial fashion. The adversarial nature of the task forces the generator to focus on the most class-discriminative feature elements. A visualization of our cross-modal feature generating method can be seen in Fig.~\ref{fig:ourMethod}.

\subsection{Multimodal Prototype}
Having learned a strong text-to-image feature mapping $G_t$ on data provided for the base classes $\Cbase$ we can employ the conditional network to generate additional visual features $G_t(\Phi_T(t_i))$ given an textual description $t_i$ and a pre-trained text encoder $\Phi_T(\cdot)$. This allows for computing a prototype from generated samples $G_t(t_i)$ according to

\begin{equation}\label{eq:protText}
    p_T^k = \frac{1}{|\Straink|}\sum_{(t_i,y_i) \in \Straink}G_t(\Phi_T(t_i)).
\end{equation}
Next, having both the true visual prototype $p^k$ from Eq.~\ref{eq:protVis} and a prototype $p_T^k$ computed from generated feature vectors conditioned on textual descriptions from Eq.~\ref{eq:protText} a new joint prototype can be computed using a weighted average of both representations:

\begin{equation}\label{eq:avg}
    p^k = \frac{p^k + \lambda * p_T^k}{1+\lambda}, 
\end{equation}
where $\lambda$ is a weighting factor and $k\in\Cnovel$ represents the class label of the prototype. Note that the step in Eq.~\ref{eq:avg} can be repeated multiple times, because $G_t$ allows for the generation of a potentially infinite number of visual feature vectors in $\varphi$.
The prediction of the class membership of unseen test samples can now be performed with Eq.~\ref{eq:nn} using the updated prototypes.
\section{Experiments}

\begin{table*}[h]
    \setlength{\tabcolsep}{10pt}
	\centering{
		\begin{tabular}{lcc} \toprule
		Method  & 1-shot & 5-shot\\ \midrule
		    MAML \cite{finn2017model}& 38.43 & 59.15\\
		    Meta-Learn LSTM \cite{ravi_optimization_2017} & 40.43 & 49.65\\
		    Matching Networks \cite{vinyals_matching_2016} & 49.34&59.31 \\
		    Prototypical Networks \cite{snell_prototypical_2017} & 45.27& 56.35\\
		    Metric-Agnostic Conditional Embeddings \cite{hilliard2018few}& 60.76&74.96 \\
		    ResNet-18 \cite{chen2018semantic}& 66.54 $\pm$ 0.53& 82.38 $\pm$ 0.43\\
		    ResNet-18 + Gaussian \cite{chen2018semantic}& 65.02 $\pm$ 0.60& 80.79 $\pm$ 0.49\\
		    ResNet-18 + Dual TriNet \cite{chen2018semantic}& 69.61 $\pm$ 0.46 & 84.10 $\pm$ 0.35\\\midrule
            Image Only Baseline (ResNet-18 + NN)& 68.85 $\pm$ 0.86 & 83.93 $\pm$ 0.57\\
            Our Full Method (Multimodal ResNet-18 + NN)& \textbf{75.01 $\pm$ 0.81} & \textbf{85.30 $\pm$ 0.54} \\\bottomrule
		\end{tabular}
	}
	\captionsetup{position=above}
	\captionof{table}{Top-1 accuracies for the 5-way classification task on the CUB-200 dataset of our approach compared with single-modal state-of-the-art few-shot learning methods. We report the average accuracy of 600 randomly samples few-shot episodes including 95\% confidence intervals.}
	\label{tab:results_singlemodal}
	\vspace{-0.5cm}

\end{table*}

To confirm the general applicability of our method we perform several experiments using two datasets. These experiments include comparisons to state-of-the-art multimodal and single-modal approaches for few-shot learning.

\subsection{Datasets}
We test our method on two fine-grained multimodal classification datasets. Specifically, we use the CUB-200-2011~\cite{WahCUB_200_2011} with bird data and Oxford-102~\cite{nilsback2008automated} containing flower data for our evaluation.
The CUB-200 dataset contains 11,788 images of 200 different bird species, with $\mathcal{I} \subset \mathbb{R}^{256\times256}$. 
The data is split equally into training and test data. 
As a consequence, samples are roughly equally distributed, with training and test set each containing $\approx 30$ images per category. 
Additionally, 10 short textual descriptions per image are provided by \cite{reed_learning_2016}. 
Similar to \cite{zhang_stackgan++:_2017}, we use the text-encoder pre-trained by Reed et al.~\cite{reed_learning_2016}, yielding a text embedding space $\mathcal T \subset \mathbb{R}^{1024}$ with a CNN-RNN-based encoding function.
Following \cite{zhang_stackgan++:_2017}, we split the data such that $\left|C_{base}\right|=150$ and $\left|C_{novel}\right|=50$. 
To simulate few-shot learning, $n\in\{1,2,5,10,20\}$ images of $C_{novel}$ are used for training, as proposed by \cite{hariharan_low-shot_2017}. We perform 50-way classification, such that during test time, all classes are considered for the classification task.
In contrast, the Oxford-102 dataset contains images of 102 different categories of flowers. Similar to the CUB-200 dataset, 10 short textual descriptions per image are available. As for the CUB-200 dataset, we use the text-encoder pre-trained by Reed et al.~\cite{reed_learning_2016}, yielding a text embedding space $\mathcal T \subset \mathbb{R}^{1024}$. Following Zhang et al.~\cite{zhang_stackgan++:_2017}, we split the data such that $\left|C_{base}\right|=82$ and $\left|C_{novel}\right|=20$. 
To simulate few-shot learning, $n\in\{1,2,5,10,20\}$ images of $C_{novel}$ are used for training. Again, we perform classification among all available novel classes, yielding a 20-way classification task.

\subsection{Implementation Details}
\paragraph{Image Encoding}
For image encoding we utilize  a slightly modified version of the ResNet-18 architecture \cite{he2016deep}. Specifically, we halve the dimensionality of every layer and add two 256-dimensional fully connected layers with LeakyRelu activation after the last pooling layer, followed by a softmax classification layer with $|\Cbase|$ units. This network is trained on base classes using the Adam optimizer \cite{kingma2014adam} for 200 iterations with learning rate $10^{-3}$, which is decreased to $5\times10^{-4}$ after 20 iterations. The last fully connected layer is employed as embedding space $\varphi$.
\paragraph{Cross-Modal Generation} For the text-to-image feature mapping we use a tcGAN architecture inspired by StackGAN++ \cite{zhang_stackgan++:_2017}. In the generator $G_t$, following \cite{zhang_stackgan++:_2017} the text embedding $\Phi_T(t_i)$ is first passed into a conditioning augmentation layer to condense the input space during training. This is followed by some upsampling convolutions, yielding a 256-dim output vector, equivalent to the dimensionality of the image feature space $\varphi$. Given the calculated feature vector $G_t(\Phi_T(t_i)$ and the original text embedding $\Phi_T(t_i)$, the discriminator $D$ outputs a conditional and an unconditional loss (see \cite{zhang_stackgan++:_2017}) along with the auxiliary classification loss. Adam is used to optimize both networks with learning rate $2\times10^{-4}$ for 500 iterations. Having  trained a feature generating network $G_t$, we compute $G_t(\Phi_T(t_i))$ for all 10 available textual descriptions per image and take the average in $\varphi$ as its feature representation.

\begin{figure*}[t!]
	\centering
  \includegraphics[width=0.7\textwidth]{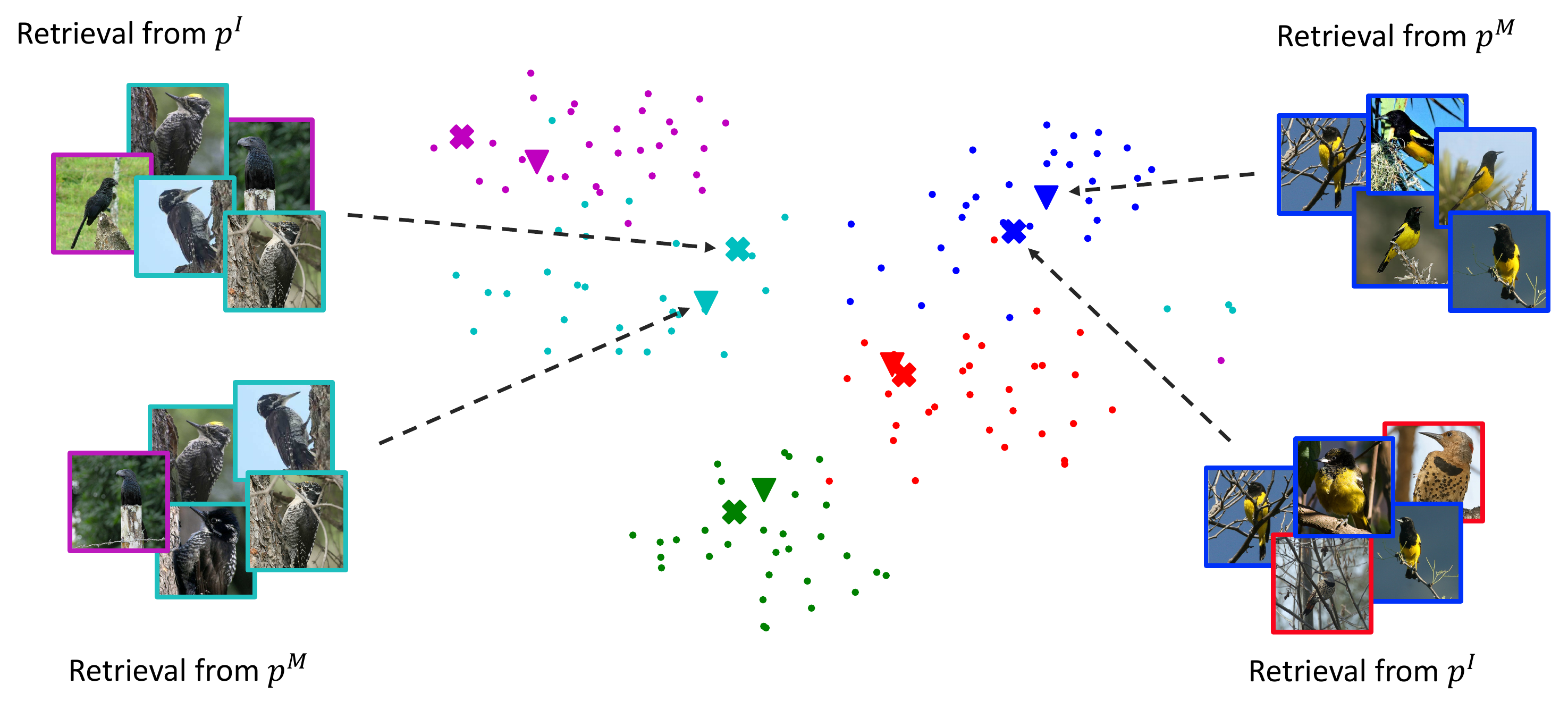}
	\caption{tSNE visualization of test samples (dots), prototypes $p_I$ computed from only real image features (crosses) and multimodal prototypes $p_M$ computed from real image features and generated features conditioned on text (triangles) in 5-way 1-shot scenario for CUB-200. The color indicates the class membership. Furthermore, we show the top-5 results for an image retrieval task for unseen images given the image-only prototype $p_I$ and the multimodal prototype $p_M$. The color of the border indicates the class membership.}
	\label{fig:tsne}
\end{figure*}

\paragraph{Classification} We predict the class membership of test samples by calculating the nearest prototype in the embedding space $\varphi$ (see Eq.~\ref{eq:nn}). As distance function we use cosine distance. To average visual and textual prototypes we set $\lambda=1$ (see Eq.~\ref{eq:avg}) and repeat this step 10 times, updating $G_t$ in every iteration. Hence, in each iteration we reuse real samples from $\Straink$, combined with novel generated samples given an updated generator $G_t$.

\subsection{Results}
For the evaluation, we test our approach in the 50-way classification task for CUB-200, and 20-way classification for Oxford-102. We designed a strong baseline, in which we predict the class label of unseen test samples by finding the nearest prototype in the the embedding space $\varphi$, where the prototype $p_I^k$ is computed exclusively using the limited visual samples (\textbf{image only}). Note that nearest neighbor classification is a powerful baseline in the context of few-shot learning, as similarly suggested by other works \cite{bauer2017discriminative}.

Furthermore, we evaluate our method in a zero-shot setting, in which we generate feature vectors given the textual descriptions.  The class-label of unseen test samples is predicted by computing the nearest prototype $p_T^k$ containing exclusively generated features conditioned on the textual descriptions (\textbf{ZSL}). Our full method calculates the average of both prototypes (\textbf{multimodal}). 

We compare our method with \cite{pahdeWACV}, which to the best of our knowledge is the only existing work leveraging multimodal data in the context of few-shot learning. 
Because the classification results highly depend on the choice of samples available in a few-shot scenarios, we run the experiments 600 times following \cite{snell_prototypical_2017} and sample a random few-shot episode, i.e. a random choice of $n$ samples per class in every iteration to cover randomness. We report the average top-5 accuracy including 95\% confidence intervals in Tab.~\ref{tab:results_multimodal}.

It can be observed that in every $n$-shot scenario we outperform our strong baselines and the other existing approach for multimodal few-shot learning. In the CUB-200 dataset, we outperform the baselines by a large margin, confirming our assumption that multimodal data in training is beneficial. For Oxford-102 the margins are lower, however, we still increase the classification results and outperform state-of-the-art results. Interestingly, our approach also stabilizes the results as the confidence intervals decrease compared to the baselines. 

\subsection{Comparison to Single-modal Methods}
Due to the lack of existing approaches leveraging multimodal data for few-shot learning, we additionally compare our approach to existing methods using only image data during training. Outperforming these state-of-the-art image-only few-shot learning proves the beneficial impact of additional text data during training. Specifically, we compare our method with MAML~\cite{finn2017model}, meta-learning LSTM~\cite{ravi_optimization_2017}, matching networks~\cite{vinyals_matching_2016}, prototypical networks~\cite{snell_prototypical_2017} and metric-agnostic conditional embeddings~\cite{hilliard2018few}. The results for CUB-200 for these methods are provided in~\cite{chen2018semantic}. We also include theirs results in our comparison. However, their experimental setup differs slightly from our evaluation protocol. Instead of performing 50-way classification, the results in~\cite{chen2018semantic} are reported for 5-way classification in the 1- and 5-shot scenarios. This implies that in every few-shot learning episode, 5 random classes are sampled for which a classification task has to be solved, followed by the choice of $n$ samples that are available per class. For the sake of comparability, we also evaluated our approach in the same experimental setup. We repeat our experiment for 600 episodes and report average top-1 accuracy and 95\% confidence intervals in Tab.~\ref{tab:results_singlemodal}.

It can be observed that even our image-only baseline, which performs nearest neighbor classification using prototypes in our modified ResNet-18 feature representation reaches state-of-the-art accuracies. Note that out image-only baseline can be interpreted as ResNet-version of Prototypical Network \cite{snell_prototypical_2017}, which uses a simpler model network architecture in its original version. Including multimodal data during training outperforms the other approaches in both 1- and 5-shot learning scenarios. This proves the strength of our nearest neighbor baseline and shows that enriching the embedding space $\varphi$ with generated features conditioned on data from other modalites further improves the classification accuracies. In Fig.~\ref{fig:tsne} we show a tSNE visualization of the embedding space $\varphi$ including the image-only and multimodal prototypes $p_I$ and $p_M$ respectively in the 5-way classification task. The graph clearly shows some clusters indicating the class membership. It can be observed that the generated feature vectors shift the prototypes into regions where more unseen test samples can be classified correctly. Moreover, Fig.~\ref{fig:tsne} shows retrieval results of unseen classes for $p_I$ and $p_M$. See further retrieval results in the Supplementary Materials.

\section{Analysis}
In order to get a further in-depth understanding of certain aspects of our method, we performed some additional experiments analyzing its behavior. To this end, we use the CUB-200 dataset for the experiments in this section.
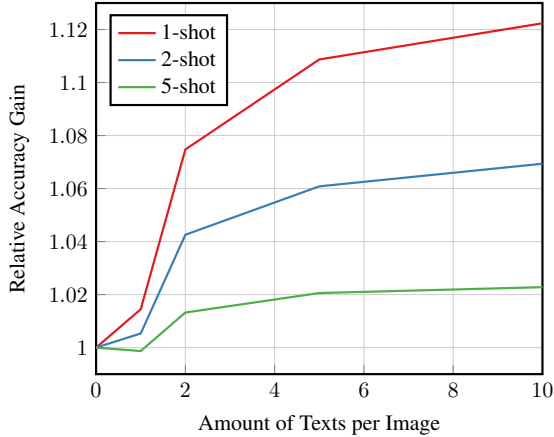
\begin{figure}[ht!]
	\centering
    \begin{tikzpicture}[scale = 0.83]%
			\begin{axis}[
            legend columns=1,
            width=0.5\textwidth,
			legend pos=north west,
            major grid style={line width=.1pt,draw=gray!40},
			xtick={0,2,4,...,10},
			grid=major,
            ytick={1, 1.02,...,1.13},
            ymin=0.99,
            xlabel={Amount of Texts per Image},
            ylabel= Relative Accuracy Gain,
            xtick pos=left,
            ytick pos=left,
            xmin=0,xmax=10,
            ymax=1.13]
            \addplot+[mark=none] table [x=iteration, y=1shot, col sep=comma] {data/reducedTexts.csv};
            \addplot+[mark=none] table [x=iteration, y=2shot, col sep=comma] {data/reducedTexts.csv};
            \addplot+[mark=none] table [x=iteration, y=5shot, col sep=comma] {data/reducedTexts.csv};
            \legend{1-shot, 2-shot, 5-shot}
            \end{axis}
		\end{tikzpicture}%
	
	\caption{Relative top-5-accuracy gain for different amounts of available texts. The y-axis shows the accuracy gain in relation to the image-only baseline and the x-axis the amount of available texts per image $k$.}
    \label{fig:reducedTexts}
\end{figure}
\subsection{Reducing Textual Data}
In a first experiment we want to analyze the importance of the amount of available textual descriptions. Note that for the experiments in Tab.~\ref{tab:results_multimodal} we used all 10 textual descriptions per image to generate a feature vector $G_t(\Phi(t_i)$. In this experiment we want to understand how the model behaves at reduced text availability. Therefore, in addition to limiting the amount of available images per novel class to $n$, we limit the amount of textual descriptions per image to $k \in \{1,2,5,10\}$. We evaluate the classification accuracy for $n \in \{1,2,5\}$ with reduced number of textual descriptions. In Fig.~\ref{fig:reducedTexts} we show the relative accuracy gains for the different amount of texts compared to the image-only baseline. The x-axis shows the amount of texts and the y-axis the relative accuracy gain. It can be observed that the lower the amount of images $n$ the higher is the accuracy gain given the text. The graphs show an increasing trend which indicates that the more texts are available the more the classification accuracies can be increased. This proves our assumption that enriching the embedding space $\varphi$ is crucial to reach high classification results. Interestingly, in every $n$-shot scenario the second text leads to the highest accuracy gain. However, adding more text constantly improves the results and is never harmful to the model.


\subsection{Impact of Prototype Shift}
We investigate how the adjustment of a certain prototype impacts the classification performance. 
Therefore, we analyze the per-class accuracy gain in correlation with the shift of the prototype when exposed to multimodal data. The assumption we want to confirm whether large adjustments to the prototype go along with higher accuracy gain compared to classes for which the prototype remains almost unchanged. To this end, we measure change in prototype  between the original image-only prototype $p_I$ and the updated multimodal prototype $p_M$ using the cosine distance denoted by $d(p_I, p_M)$. For every novel class, we analyze the correlation of the prototype update to the accuracy gain compared to the image-only baseline.
\begin{figure}[ht!]
	\centering
    \begin{tikzpicture}[scale = 0.83]
    \begin{axis}[
            ybar,
            width=0.5\textwidth,
            bar width=0.7,
            xtick={0,10,20,...,50},
            ytick={0,0.05,0.1,0.15},
            xlabel={Prototypes sorted by $d(p_I, p_M)$},
            ylabel=Per-Class Accuracy Gain,
            xtick pos=left,
            ytick pos=left,
            grid=major,
            scaled ticks=false, tick label style={/pgf/number format/fixed} 
        ]
        \addplot+table[x=iteration,y=acc, col sep=comma] {data/rankingResults.csv};
    \end{axis}
\end{tikzpicture}
		\caption{Per-class accuracy gain for prototypes after the adjustment with generated feature vectors. The x-axis shows the rank of the prototype sorted by $d(p_I, p_M)$ and the y-axis the top-5-accuracy gain for that particular class.}
    \label{fig:ranking}
\end{figure}
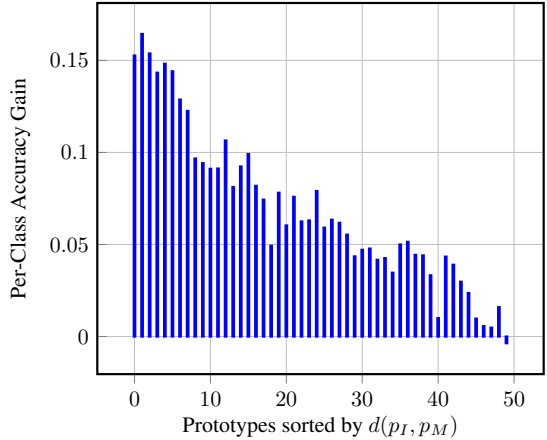
In Fig.~\ref{fig:ranking} we show the per-class accuracy gain for all prototypes in the 1-shot scenario. The x-axis shows the rank of the prototypes for all 50 novel classes of the CUB-200 dataset sorted by $d(p_M, p_I)$ in a descending order. The y-axis represents the accuracy gain for the certain prototype. We report top-5 accuracy and show the average of the result for 100 few-shot episodes. It can be observed that the more the prototype is changed (low rank) the higher is the accuracy gain for this particular class. On average, the most changed prototype leads to a per-class top-5 accuracy gain of ca. 15\%. Smaller changes have a smaller impact on the classification performance and on average, adjusting the prototype with multimodal data is not harmful for the accuracy. This suggests that the multimodal features carry complimentary information that is used to simulate unseen novel class samples. At the same time it shows that the text-to-image feature mapping is well learned, as the most diverse, or farthest multimodal features net the largest performance gains.


\section{Conclusion and Future Work}

In this paper we tackled the few-shot learning problem from a multimodal perspective. Therefore, we proposed to leverage a nearest neighbor classifier in a powerful representation space. To mitigate the low population problem caused by the few-shot scenario we developed a cross-modal generation framework that is capable of enriching the visual feature space given data in another modality (e.g. text). Classification can now be performed by finding the nearest multimodal class prototype to an unseen test sample. We evaluated our proposed methods on the two multimodal datasets CUB-200 and Oxford-102 and showed the applicability of our approach. We outperformed our strong baselines and state-of-the-art single-modal and multimodal methods by a large margin. 
For future work we plan to follow the notion of \cite{schonfeld2019generalized} which entails optimizing jointly the learned representation on multimodal data.


\FloatBarrier
{\small
\bibliographystyle{ieee}
\bibliography{wacv2021}
}

\clearpage
\onecolumn
\begin{appendices}


    \section{Extended Quantitative Results}
\paragraph{Results for multimodal 50-way classification}
We extend Tab. 1 from the main paper which shows the accuracies for 50-way classification in comparison with state-of-the-art multimodal few-shot learning methods and our baselines. While in the paper we only reported the top-5 accuracy, we provide top-1 and top-3 accuracy in addition in Tab.~\ref{tab:results_multimodal}. We can observe the same trends as for the top-5 accuracy. 

\begin{table*}[ht]
    \setlength{\tabcolsep}{5pt}
	\centering{
		\begin{tabular}{lccccccc} \toprule
		Dataset & Method & Metric & 1-shot & 2-shot & 5-shot & 10-shot &20-shot \\ \midrule
		CUB & Pahde et al. \cite{pahdeWACV} & Top-1& 24.90 & 25.17 & 34.66 & 44.00 & 53.70 \\
		&&Top-3& 37.59 & 39.75 & 49.86 & 59.62 & 67.99 \\
		&&Top-5& 57.67 & 59.83 & 73.01 & 78.10 & 84.24 \\
		& Image Only Baseline & Top-1&28.91$\pm$0.18 & 37.52$\pm$0.15 & 47.33$\pm$0.12 & 52.31$\pm$0.12 & \textbf{55.62$\pm$0.14}\\
		&&Top-3& 51.13$\pm$0.22 & 62.23$\pm$0.16 & 72.53$\pm$0.11 & 76.71$\pm$0.09 & 78.87$\pm$0.10\\
		&&Top-5& 62.65$\pm$0.22 & 73.52$\pm$0.15 & 82.44$\pm$0.09 & 85.64$\pm$0.08 & 87.27$\pm$0.08 \\
		& ZSL Baseline  & Top-1& 22.39$\pm$0.18 & 27.15$\pm$0.16 & 32.24$\pm$0.15 &  34.65$\pm$0.13 & 36.42$\pm$0.17\\
		&&Top-3& 45.37$\pm$0.21 & 52.42$\pm$0.18 & 59.09$\pm$0.14 & 61.73$\pm$0.13 & 63.27$\pm$0.15 \\
		&&Top-5& 58.28$\pm$0.22 & 65.62$\pm$0.19 & 71.79$\pm$0.14 & 74.15$\pm$0.11 & 75.32$\pm$0.13 \\
		& Our (Multimodal) &Top-1&  \textbf{34.16$\pm$0.17} & \textbf{41.43$\pm$0.14} & \textbf{48.84$\pm$0.10} & \textbf{53.01$\pm$0.11} & 55.58$\pm$0.14\\
		&&Top-3& \textbf{58.56$\pm$0.19} & \textbf{67.44$\pm$0.13} & \textbf{74.65$\pm$0.09} & \textbf{77.60$\pm$0.09} &  \textbf{79.30$\pm$0.11}\\
		&&Top-5& \textbf{70.39$\pm$0.19} & \textbf{78.62$\pm$0.12} & \textbf{84.32$\pm$0.06} & \textbf{86.23$\pm$0.08} & \textbf{87.47$\pm$0.09} \\\midrule
		Oxford-102 & Pahde et al. \cite{pahdeWACV}
		&Top-1& 43.77 &	61.42	& 72.49 & - & - \\
		&&Top-3& 57.96	&77.68	&82.18 & - & - \\
		&&Top-5& 78.37 & 91.18 & 92.21 & - & - \\

		& Image Only Baseline & Top-1&  49.39$\pm$0.34 &60.02$\pm$0.27 & 70.24$\pm$0.18 & 74.49$\pm$0.16 & \textbf{76.98$\pm$0.16}\\
		&&Top-3& 74.12$\pm$0.32  & 82.76$\pm$0.21&  89.11$\pm$0.12& 91.34$\pm$0.11 & 92.46$\pm$0.10\\
		&&Top-5& 83.88$\pm$0.27 &  90.27$\pm$0.16 & 94.25$\pm$0.09 & 95.61$\pm$0.08 & 96.21$\pm$0.08 \\

		& ZSL Baseline  & Top-1& 34.99$\pm$0.32  & 39.63$\pm$0.29& 42.92$\pm$0.24 & 44.21$\pm$0.22 & 44.91$\pm$0.21\\
		&&Top-3& 60.90$\pm$0.34 &65.91$\pm$0.30 & 69.89$\pm$0.23 & 71.70$\pm$0.21 & 72.76$\pm$0.20 \\
		&&Top-5& 73.14$\pm$0.31 & 77.63$\pm$0.27 & 81.10$\pm$0.19 & 82.88$\pm$0.18 & 84.00$\pm$0.17 \\

		& Our (Multimodal) &Top-1& \textbf{53.70$\pm$0.30}   & \textbf{62.46$\pm$0.24}& \textbf{71.08$\pm$0.19} & \textbf{74.73$\pm$0.17} &76.77$\pm$0.15\\
		&&Top-3& \textbf{77.71$\pm$0.25}  &\textbf{84.35$\pm$0.18} & \textbf{89.51$\pm$0.11} & \textbf{91.53$\pm$0.11} & \textbf{92.48$\pm$0.10}\\
		&&Top-5& \textbf{86.34$\pm$0.21}& \textbf{91.30$\pm$0.15} & \textbf{94.58$\pm$0.09} & \textbf{95.77$\pm$0.08} & \textbf{96.34$\pm$0.07}\\
		\bottomrule
		\end{tabular}
	}
	\captionsetup{position=above}
	\captionof{table}{50-way classification top-1, top-3 and top-5 accuracies in comparison to other multimodal few-shot learning approaches and our baselines for CUB-200 and Oxford-102 datasets with $n\in\{1,2,5,10,20\}$. The best results are in bold.}
	\label{tab:results_multimodal}
\vspace{-0.5cm}
\end{table*}

\paragraph{Results for additional k-way n-shot classification tasks}
To extend the results in Tab. 2 in the main paper, we evaluate our method in additional k-way classification scenarios with $k \in \{5,10,20\}$. For this experiment we compare our full method to the image-only baseline. Further extending Tab. 2 from the main paper, we report top-3 and top-5 accuracy in addition to the top-1 accuracy. The results for different $n$-shot scenarios with $n \in \{1,2,5,10,20\}$ are shown in Tab.~\ref{tab:results_singlemodal}. It can be observed that in the most scenarios our method improves the classification accuracies and decreases the variance. However, the more visual data is available (higher $n$), the less impact have the generated features.
\begin{table*}[h!]
    \setlength{\tabcolsep}{6pt}
	\centering{
		\begin{tabular}{llcccccc} \toprule
		Task & Method  & Metric & 1-shot & 2-shot & 5-shot & 10-shot & 20-shot\\ \midrule
            5-way & Image Only Baseline& Top-1 & 68.85$\pm$0.86 & 79.99$\pm$0.66 & 83.93$\pm$0.57& 86.95$\pm$0.49 &\textbf{87.78$\pm$0.48}\\
            
            && Top-3 & 94.71$\pm$0.34 &97.80$\pm$0.16 & 98.39$\pm$0.12 &98.91$\pm$0.09&98.95$\pm$0.09\\
            && Top-5 & - & - & - &-&-\\
            
            & Our (Multimodal)&Top-1 &  \textbf{75.01 $\pm$ 0.81} & \textbf{80.90$\pm$0.64}& \textbf{85.30 $\pm$ 0.54} & \textbf{86.96$\pm$0.48}&87.67$\pm$0.51
            \\
                        && Top-3 & \textbf{96.83$\pm$0.25} & \textbf{98.06$\pm$0.14}& \textbf{98.65$\pm$0.11} & \textbf{98.94$\pm$0.08}&\textbf{99.02$\pm$0.09}\\
            && Top-5 & - &- & - &-&-\\\midrule

            10-way & Image Only Baseline& Top-1 & 59.34$\pm$0.76 & 68.25$\pm$0.65 & 75.72$\pm$0.61 &78.12$\pm$0.60&\textbf{79.68$\pm$0.85}\\
                        && Top-3 & 86.19$\pm$0.51 & 91.67$\pm$0.33 & 94.66$\pm$0.25& \textbf{95.65$\pm$0.24} &\textbf{95.99$\pm$0.31}\\
            && Top-5 & 94.50$\pm$0.29 & 97.20$\pm$0.15 & 98.24$\pm$0.11& 98.55$\pm$0.11 &\textbf{98.63$\pm$0.14}\\
            
            & Our (Multimodal)&Top-1 & \textbf{62.25$\pm$0.73}  & \textbf{69.71$\pm$0.63}  & \textbf{76.02}$\pm$0.62& 78.12$\pm$0.64 &79.29$\pm$0.82\\
                        && Top-3 & \textbf{88.41$\pm$0.45} & \textbf{92.78$\pm$0.28} & \textbf{94.88$\pm$0.23}& 95.55$\pm$0.25 &95.89$\pm$0.31\\
            && Top-5 & \textbf{95.80$\pm$0.25} & \textbf{97.57$\pm$0.14} & \textbf{98.32$\pm$0.11} & \textbf{98.57$\pm$0.11} &98.62$\pm$0.12\\
            \midrule

            20-way & Image Only Baseline& Top-1 & 46.55$\pm$0.40 & 55.94$\pm$0.36& 64.64$\pm$0.39 & \textbf{68.21$\pm$0.43} &\textbf{69.97$\pm$0.47}\\
                        && Top-3 & 73.60$\pm$0.38 & 81.94$\pm$0.28 & 87.93$\pm$0.23 &89.66$\pm$0.22&\textbf{90.55$\pm$0.26}\\
            && Top-5 & 84.37$\pm$0.30 & 90.56$\pm$0.19 & 94.06$\pm$0.14 &94.96$\pm$0.14&\textbf{95.49$\pm$0.15}\\
            
            & Our (Multimodal)&Top-1 & \textbf{48.23$\pm$0.40 } &\textbf{57.02$\pm$0.36}& \textbf{64.94$\pm$0.38} &68.08$\pm$0.43&69.62$\pm$0.44\\
                        && Top-3 &\textbf{75.31$\pm$0.36} & \textbf{82.90$\pm$0.27}& \textbf{88.31$\pm$0.22}&\textbf{89.75$\pm$0.21}&90.51$\pm$0.24\\
            && Top-5 & \textbf{85.65$\pm$0.29} & \textbf{91.23$\pm$0.18} & \textbf{94.27$\pm$0.14}&\textbf{95.02$\pm$0.13}&95.42$\pm$0.14 \\
            \bottomrule
		\end{tabular}
	}
	\captionsetup{position=above}
	\captionof{table}{Top-1, top-3 and top-5 accuracies for different $k$-way classification tasks on the CUB-200 dataset of our approach compared to our image-only baseline. We report the average accuracy of 600 randomly sampled few-shot episodes including 95\% confidence intervals. The best results are in bold.}
	\label{tab:results_singlemodal}

\end{table*}

    \FloatBarrier
\section{Visualization of Embedding Space}
\begin{figure}
\centering
\begin{minipage}{.5\textwidth}
  \centering
  \captionsetup{margin=.4cm}
  \includegraphics[width=\linewidth]{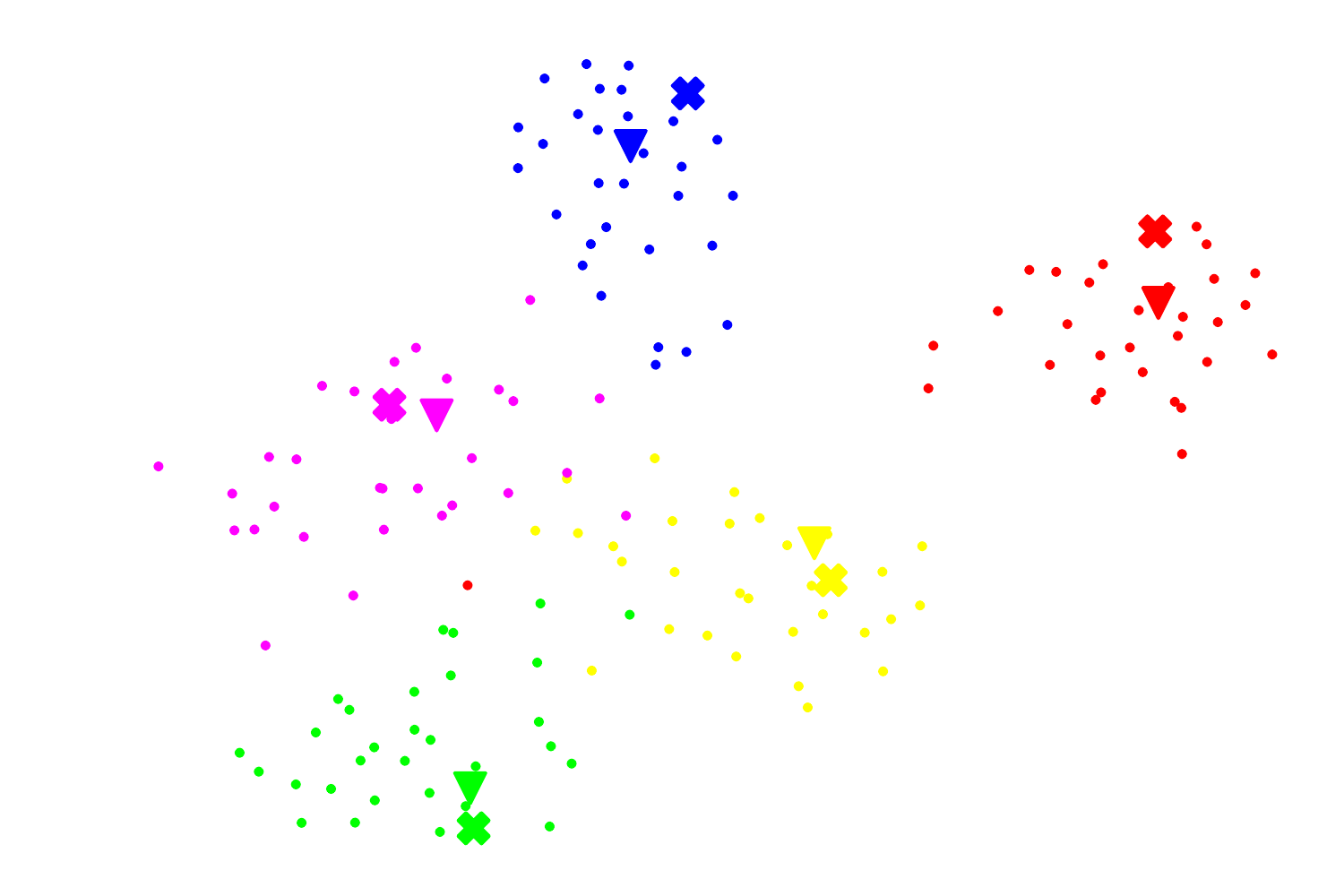}
  \captionof{figure}{t-SNE graph for 5-way classification showing the image-only prototypes (crosses), updated multimodal prototypes (triangles) and unseen test samples (dots)}
  \label{fig:tsne1}
\end{minipage}%
\begin{minipage}{.5\textwidth}
  \centering
  \includegraphics[width=\linewidth]{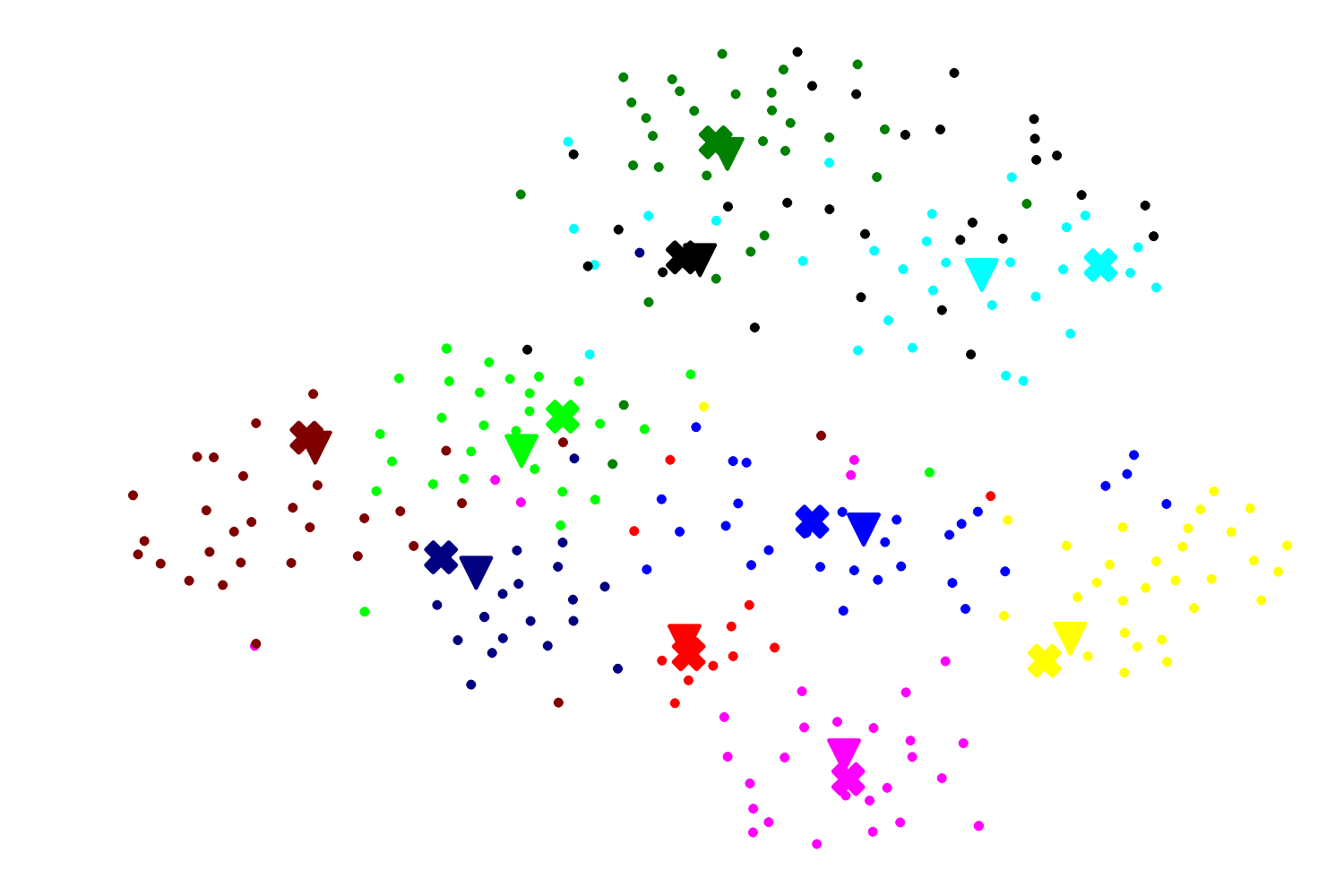}
  \captionsetup{margin=.4cm}
  \captionof{figure}{t-SNE graph for 10-way classification showing the image-only prototypes (crosses), updated multimodal prototypes (triangles) and unseen test samples (dots)}
  \label{fig:tsne2}
  
\end{minipage}
\vspace{1cm}
\end{figure}

\begin{figure}
\centering
\begin{minipage}{.5\textwidth}
  \centering
  \captionsetup{margin=.4cm}
  \includegraphics[width=\linewidth]{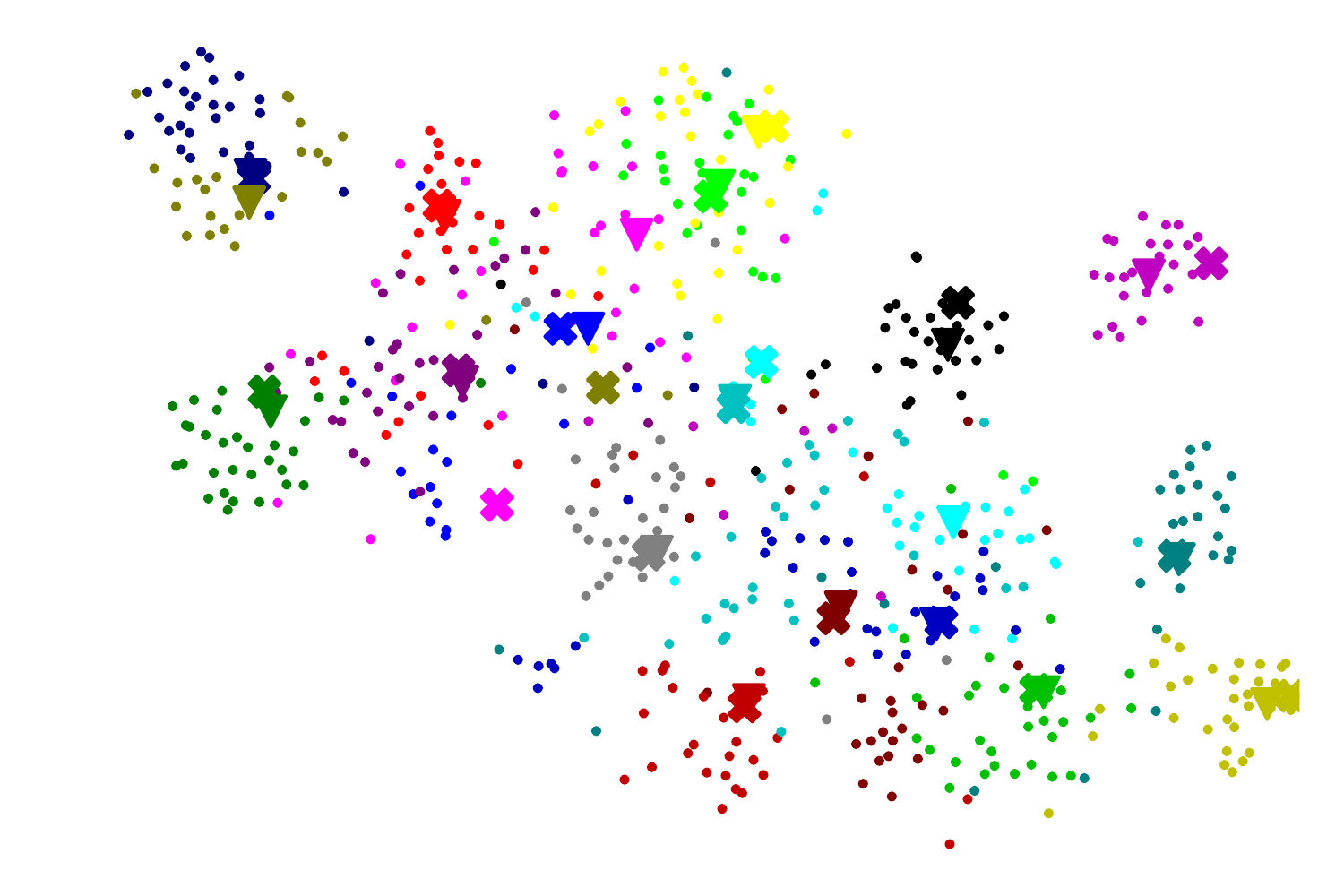}
  \captionof{figure}{t-SNE graph for 20-way classification showing the image-only prototypes (crosses), updated multimodal prototypes (triangles) and unseen test samples (dots)}
  \label{fig:tsne3}
\end{minipage}%
\begin{minipage}{.5\textwidth}
  \centering
  \captionsetup{margin=.4cm}
  \includegraphics[width=\linewidth]{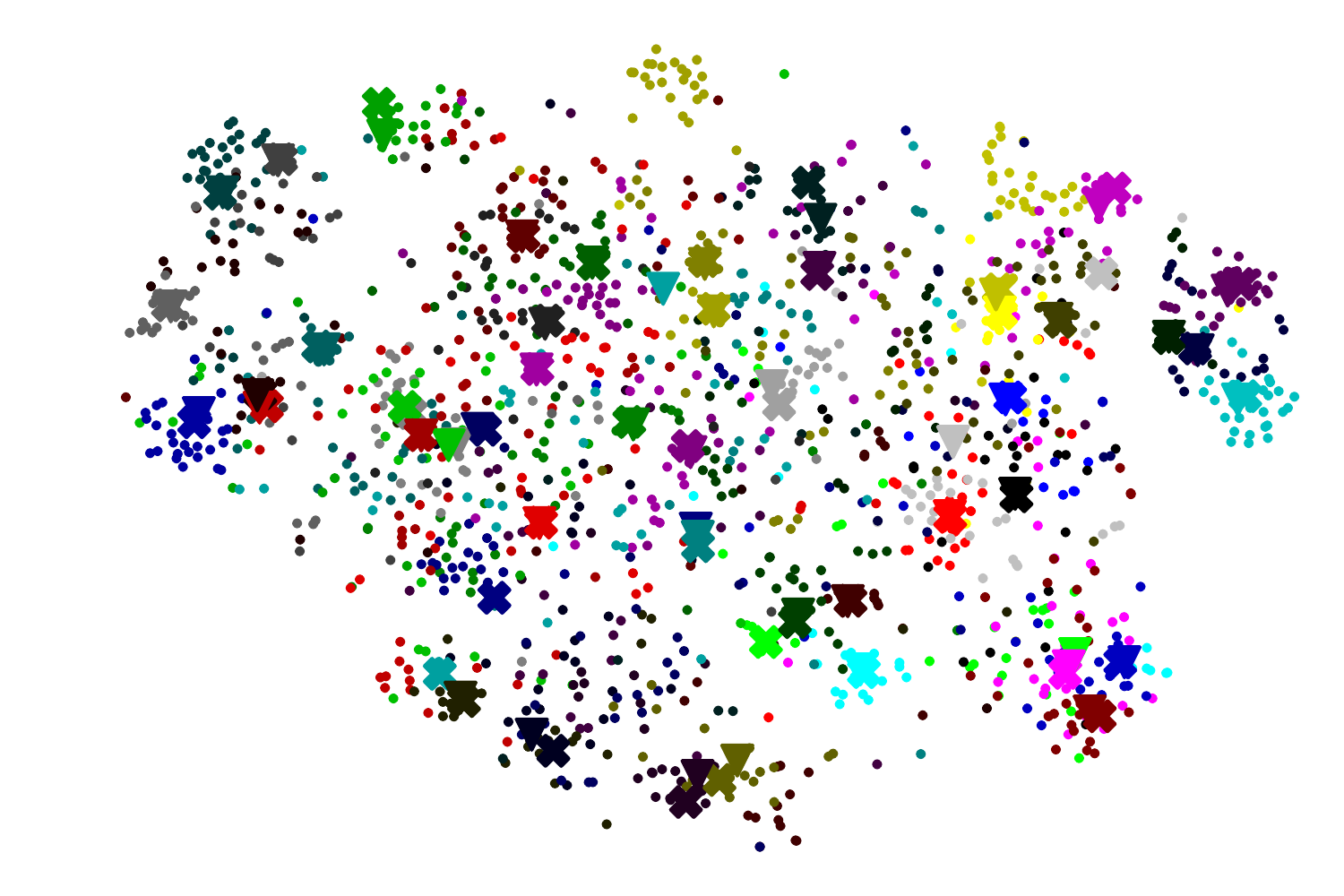}
  \captionof{figure}{t-SNE graph for 50-way classification showing the image-only prototypes (crosses), updated multimodal prototypes (triangles) and unseen test samples (dots)}
  \label{fig:tsne4}
\end{minipage}
\vspace{0.5cm}
\end{figure}

In this section we provide additional visualizations for the embedding spaces in different few-shot learning scenarios. Therefore, we show visualizations for the embedding space in $k$-way classification problems with $k \in \{5,10,20,50\}$. We use t-SNE for dimensionality reduction. The visualizations are shown in Figures~\ref{fig:tsne1}, \ref{fig:tsne2}, \ref{fig:tsne3} and \ref{fig:tsne4} in which the different colors indicate the class membership. Although the dimensionality is reduced to two for being able to visualize the embedding space we can still see the clusters for different classes. It can be observed that in many cases the multimodal prototypes (triangles) are moved towards the center of the cluster for the particular class compared to the image-only prototypes (crosses).

\FloatBarrier
\section{Retrieval Results}
\begin{figure}[t!]
	\centering
  \includegraphics[width=\textwidth]{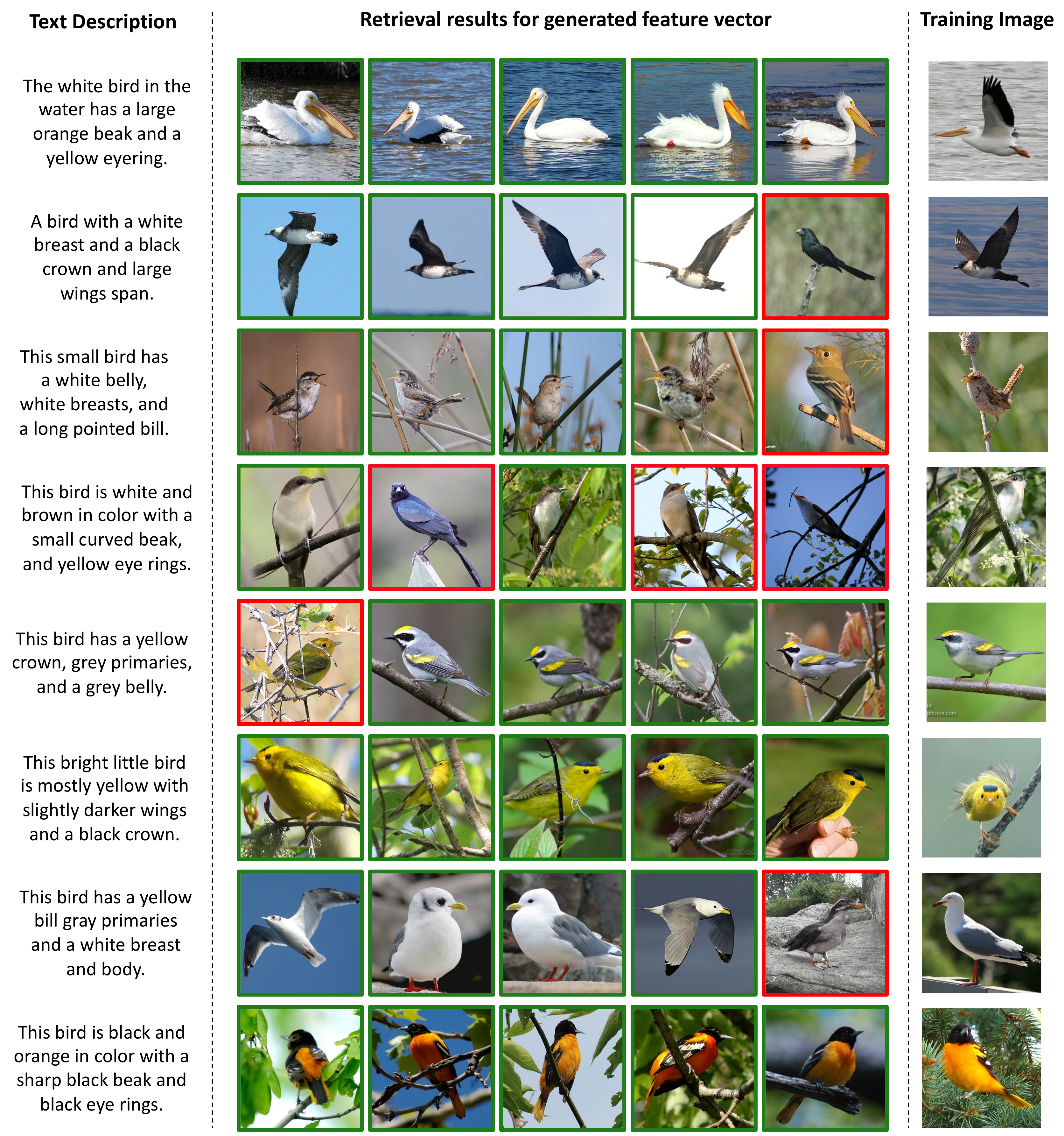}
	\caption{Retrieval Results for generated feature vectors in a 5-way classification task: The left column shows the textual description (one random caption out of the ten available descriptions per training image), in the middle are the top-5 retrieved unseen test images and in the right column is the training image for the particular class. The color of the surrounding box indicates whether the retrieved test image is from the correct class (green) or a wrong class (red).}
	\label{fig:retrieval}
\end{figure}

As an additional experiment we use the generated feature vectors to retrieve the closest unseen test images in the 5-way 1-shot learning scenario. The feature vectors are generated conditioned on the textual descriptions of the training images. We calculate the distance between feature vectors using the cosine distance measure. The retrieval results for some randomly selected feature vectors are shown in Fig.~\ref{fig:retrieval}. This shows the effectiveness of the text-conditional feature generator. It can be seen that most of the retrieved images are from the correct class. Thus, the generated feature vector is close to unseen test images of the same class, facilitating classification with a nearest neighbor approach.

\FloatBarrier
\end{appendices}

\end{document}